**IEEE** Access

Multidisciplinary : Rapid Review : Open Access Journal

# Incremental Deep Learning for Robust Object Detection in Unknown Cluttered Environments

**Dong Kyun Shin[1], Minhaz Uddin Ahmed[1] and Phill Kyu Rhee[1]**
[1]Computer Engineering Department, Inha University, Incheon, South Korea

Corresponding author: Phill Kyu Rhee (e-mail: pkrhee@inha.ac.kr)

This work was supported by Inha University research grant.

**ABSTRACT** Object detection in streaming images is a major step in different detection-based applications, such as object tracking, action recognition, robot navigation, and visual surveillance applications. In most cases, image quality is noisy and biased, and as a result, the data distributions are disturbed and imbalanced. Most object detection approaches, such as the faster region-based convolutional neural network (Faster RCNN), Single Shot Multibox Detector with 300x300 inputs (SSD300), and You Only Look Once version 2 (YOLOv2), rely on simple sampling without considering distortions and noise under real-world changing environments, despite poor object labeling. In this paper, we propose an Incremental active semi-supervised learning (IASSL) technology for unseen object detection. It combines batch-based active learning (AL) and bin-based semi-supervised learning (SSL) to leverage the strong points of AL's exploration and SSL's exploitation capabilities. A collaborative sampling method is also adopted to measure the uncertainty and diversity of AL and the confidence in SSL. Batch-based AL allows us to select more informative, confident, and representative samples with low cost. Bin-based SSL divides streaming image samples into several bins, and each bin repeatedly transfers the discriminative knowledge of convolutional neural network (CNN) deep learning to the next bin until the performance criterion is reached. IASSL can overcome noisy and biased labels in unknown, cluttered data distributions. We obtain superior performance, compared to state-of-the-art technologies such as Faster RCNN, SSD300, and YOLOv2.

## I. INTRODUCTION

Though there are great advancements in object detection technologies [1]–[3], to classify and localize visual scene objects, however, the quality of new streaming samples still poses a challenging problem in video object detection. Junwei *et al.* reviewed the recent progress on variety of object detection research work with state-of-the-art methods, benchmark dataset and their evaluation through comparing result [4]. The state-of-the-art object detection technologies, such as the fast region-based convolutional neural network (Fast RCNN) [5], OverFeat [6], Faster RCNN [7], Spatial Pyramid Pooling [8], Single Shot MultiBox Detector [9], You Only Look Once (YOLO) [10], YOLOv2 [11], YOLOv3 [12], the region-based fully convolutional network [13] and RetinaNet [14], use high-dimensional deep feature spaces and find performance degradation in detecting streaming data due to the poor quality of training samples, compared with diverse changing real-world environments. Recently, CNNs were applied successfully in the object detection/recognition area after Krischesky *et al*. in 2012 broke the performance barrier of object detection in the ImageNet competition [15]. Advanced performance of object detection using CNN technology mainly depends on the availability of large, correctly labeled datasets for training [16], [17]. Gong *et al*. proposed discriminative CNN (D-CNN) for remote sensing scene classification [18]. In order to minimize the classification error metric learning has applied to CNN features. Previous object detection schemes [5]–[14] were designed on the assumption that labeled training data samples are randomly and independently distributed. Such an assumption is not valid in real-world, streaming object–detection applications, such as autonomous driving [19], visual surveillance [20], action recognition [21]–[23], and service robotics [24]. Junwei *et al*. proposed a two-stage co-segmentation framework where union background is applied to reduce disturbing image background [25]. Underlying distributions of streaming samples are substantially imbalanced, and the collected samples are very often biased or badly labeled. Supervised learning–based detection methods adopted most of the state-of-the-art techniques, and showed the only promise for automatic object detection under the assumption that the training and testing sets have the same data distribution. Most previous object detection approaches [11], [13], [14] rely on simple sampling strategies without considering distortion and noise in the data due to changing environments. Noisy sample selection and poor-quality labeling cause imbalanced data distribution and significant performance degradation when a conventional CNN is used. However, in a real-world situation, the operational environment of an individual system widely changes. Therefore, the static world assumption is not valid in constructing an efficient object detector.

The active learning (AL) method adds more informative labeled data to the training set in each iteration from unlabeled or imperfect streaming samples. It selects the most uncertain samples (i.e., the highest disagreement among classifiers) relying on uncertainty and diversity criteria. It uses a selective sampling strategy followed by queries for continuous, adaptive, and incrementally improved detection performance. The assumption of dependable human labeling is not valid for streaming object detection. The cost of a high-quality labeling process becomes too expensive to be acceptable in real-time object detection, since strict labeling rules cannot apply in AL. Labeling is performed by a current classification model with an assumption that samples in the same cluster are likely to be of the same class. Due to wrong clustering assumptions in many real-world environments, semi-supervised learning (SSL) alone cannot produce better performance, and even degrades classification accuracy in streaming object detection. Stream-based sampling [26], adaptive sampling [27], [28], and many other approaches are often bordered by sample selection bias, where some samples are error-prone and may not satisfy a random sampling assumption to avoid local overfitting. Furthermore, in many applications, the collected dataset tends to be imbalanced in class distribution. Streaming samples used in an SSL approach result in incorrect modeling, and lead to degradation of system performance. Owing to the intrinsic complexity of fallible labeling and imbalanced data samples in object detection, we need a novel approach to sample selection and an adaptive learning method.

Researchers in the machine-learning community employ the collaborative strategy of exploring and exploiting noisy or unlabeled samples using the discrimination capability of both humans and classifiers to label the samples [29]. Such collaborative sampling methods combined with AL and SSL provide a successful direction. However, to the best of our knowledge few researchers in object detection have focused on this promising direction.

AL and SSL methods work well on classification accuracy improvement using both labeled and unlabeled data [30]–[32]. We adopted them to overcome the limitations of object detectors under dynamically varying environments. Batch mode AL is employed for learning a whole group of samples at one time, rather than learning one sample at a time. A confident sample is defined as having correct information on object category, attributes, position, and size (to be used in efficient learning), whereas a noisy sample does not. For example, the attributes of a noisy region of interest (a ground-truth object bounding box) represented by a CNN deep feature may be ill-posed or too noisy to be modeled in the training time. The dataset of noisy samples is biased and tends to be imbalanced in distribution. A noisy sample should be handled differently from confident samples, since imperfect samples do not contribute to, or might even degrade, detector performance. We first build an initial object-detection CNN model using a small number of confidently labeled samples. The noisy samples are partitioned into several clusters using a k-means algorithm and make the batch pool of informative and representative samples for AL-based exploration. Finally, we construct an adaptive object detector based on bin-based SSL, where the bin is generated with a streaming image sequence in a real-world application. It begins with the batch-based CNN model, and improves detection accuracy by increasing the confident samples using bin-based Incremental active semi-supervised learning (IASSL) in noisy streaming sample distribution. In this way, IASSL provides the means for both exploration and exploitation using combined informative and reliable sampling methods. The novelties of the proposed IASSL are summarized in the following.

1) A novel framework is proposed which significantly improve the performance of object detection by combining batch-based AL and bin-based SSL incrementally. Thus, IASSL takes advantage of both informative and reliable sampling properties. To the best of our knowledge, our work is the first research which incorporates the collaborative sampling and active semi-supervised learning in the area of object detection.

2) In real world scenario an individual system widely changes due to the noisy factors as a result the static world assumption is not suitable for efficient object detection. Our hierarchical object detection tree using IASSL partly solve this problem. Our method effectively leverages the discriminative capability of deep features with a collaborative sample-selection strategy satisfying uncertainty, diversity, and confidence requirements of both AL and the SSL methods.

3) A significant amount of experiment conducted on openly available benchmark dataset such as PASCAL VOC and MS COCO combined with local dataset. Our method obtains higher mean average precision improvement and reduced the error object–detection rate.

**IEEE** Access

The rest of this paper is organized as follows. We introduce related work in active learning, semi-supervised learning, and the combination of AL and SSL in Section II. In Section III, we propose the IASSL framework. Hierarchical object detection tree is discussed in Section IV, and Incremental active semi-supervised learning is discussed in Section V. The experimental results are given in Section VI. Finally, the concluding remarks are drawn in Section VII.

## II. RELATED WORK

### A. ACTIVE LEARNING

Active learning leverages the known information of the testing data from human decisions, which is most beneficial to the learning process of a classifier, by selecting samples that are expected to improve the learning performance of the classifier [29], [33]. Many researchers have investigated the active leaning method for image classification [30]. Active learning has been employed for region labeling and image recognition tasks with less annotation effort and better annotation quality. Since annotation quality varies from people to people, some researchers have investigated automatic annotation precision assurance [30], [31]. Assuming that a single prominent object of interest exists in an image, some approaches have tried to learn object models directly from noisy keyword search. Diverse studies have been performed in active learning frameworks such as real-time stream-based sampling and adaptive sampling.

For most active learning, only uncertain and erroneous portions of the training data are required to make queries that are annotated manually with minimum human effort. These annotated data are added to the training dataset to achieve the highest gain in classification accuracy. Therefore, the query/sampling strategy is the main consideration in the active learning technique to reduce user involvement. In 2017, Gordon *et al.* adapted an uncertainty sampling strategy to select samples closer to the decision boundary [34], [35]. While uncertainty-based sampling has the advantage of exploring uncertain boundaries to accelerate quick convergence of learning curves, it also has some disadvantages, such as selected sample points that easily include mistaken outliers. Since many uncertain sample points are often not good representatives of the whole distribution, it degrades sampling quality as well as the performance of the classifier. Settles and Craven [36] considered both informative and representative characteristics of the data distribution in their information density framework. They developed a classifier on representative cluster sets where the most representative samples were selected for label propagation in the same cluster. The sampling strategy called the query by committee (QBC) ensemble learning method relies on different hypotheses of a committee of classifiers by selecting informative samples where disagreement between the classifiers is maximal. It is a critical issue for QBC to construct an accurate and, at the same time, diverse committee.

Considering this diversity, the selection strategy of sample batches in each iteration allows speeding up the learning process with different considerations, such as minimizing the margin and maximizing diversity [33]. Batch AL was introduced to consider unlabeled examples as an optimization of the discriminative classifiers. The use of clustering in batch AL has been shown to improve the diversity of the sample selection [32], especially the clustering structure to avoid uncertain sampling redundancy. The batch of samples is determined by the Fisher criteria, considering a tradeoff between uncertainty and diversity. Monte Carlo simulation [27] is employed to select the best-matched sample distribution from a sequential policy and to query the samples. In 2017, Ucar *et al.* introduced semi-supervised learning that leveraged their classifier performance using both labeled and unlabeled samples, which was applied when huge unlabeled or imperfectly labeled data are available [19]. However, the amount of labeled data used in their learning method was relatively small. SSL uses unlabeled or noisy data directly in the training process without any human-labeling efforts [28], [30], [31].

SSL approaches are divided into self-training, co-training, generative probabilistic models, and graph-based SSL. In self-training SSL, the classifier is first constructed with a small amount of labeled training samples, and secondly, a portion of the unlabeled training dataset is labeled using the current classifier. Thirdly, the most confident samples among the predicted labeled samples are repeatedly added to the training dataset until it achieves convergence. The uncertainty sampling method in AL is a complementary approach, where the least confident samples are selected for querying. In co-training, an ensemble method is employed. First, separate models are learned using independently labeled datasets. The current models classify the unlabeled data, and learn the next models using a few selected samples with the most confident predicted labels, which minimizes the version space.

### B. COMBINATION OF AL AND SSL

AL and SSL try to solve the same problem from opposite points of view. AL and SSL are based on different principles but have the common goal of high classification accuracy with minimum human-labeling effort [30]. AL and SSL methods can be combined to exploit both labeled and tentatively labeled samples for classifier training and to explore new samples labeled manually. Combination methods of AL and SSL are divided into sequential combination, SSL embedding in AL, and collaborative labeling. Sequential combination understands that the initial training set is critical for SSL to converge to the target performance. This method employs AL in order to establish an effective initial training set in the first phase, and it allows SSL to improve accuracy by using the unlabeled samples in the second phase. First, AL is applied iteratively to add additional labeled samples to constitute informative training samples, and SSL improves the classification accuracy by leveraging the information of unlabeled samples. For example, QBC is combined with Expectation-Maximization SSL in a sequential manner and is employed to assign labels to unlabeled samples.

SSL embedding in AL treats SSL as the classifier in an AL framework. Muslea *et al*. adopted this strategy by employing multiple views for both AL and SSL [32]. Exploiting unlabeled data in another manner allows both human experts and classifiers to collaboratively label the selected noisy or unlabeled data. In each iteration, unlabeled samples are chosen with a sample selection strategy and labeled by the criteria of human experts for uncertainty and by the current classifier for high classification confidence. Human effort can be much reduced in order to expand the labeled dataset to satisfy the contradictory requirements of uncertainty and confidence criteria.

The AL and SSL combining method is based on several different architectures [28], [30]. Lunjun Wan *et al*. performed AL based verification for low confidence pseudo-labeled samples labeled by SSL [37]. Manually labeled samples are dealt as candidate samples for training of the classifier in every iteration of the collaborative labeling approach. Collaboratively combining AL and SSL using the confidence score from a boosting algorithm was applied to a spoken language classification problem [38]. Several other ways of combining AL and SSL were studied [28], [32]. Tuia *et al*. proposed a large-scale image classifier, whereby a classifier is built on the hierarchical clustering model, and unlabeled pixels are selected with higher classification uncertainty [39]. Clusters are recursively split into child clusters until each leaf cluster is assigned to a class to minimize classification error, where several active query strategies are applied for selecting most uncertain samples from among the clusters. Multi-view disagreement–based AL is combined with manifold-based SSL to select highly representative samples [30]. The multi-view approach constitutes a contention pool by selecting unlabeled samples with the highest disagreement, and learns the correct class by minimizing the disagreement among the different views. A manifold learning technique [30] was applied to the samples in the contention pool and their labeled and unlabeled adjacent samples, where the samples with the largest inconsistency are chosen and labeled manually. In some approaches [29], [30], SSL assists AL in deciding the most informative label. After the increment of the labeled dataset by the AL process, a classifier is constructed by applying an SSL algorithm to the labeled and the remaining unlabeled samples.

## III. SYSTEM OVERVIEW

We introduce an efficient collaborative sampling strategy, IASSL in Fig. 1, to alleviate selection bias and the class distribution imbalance that can occur frequently in streaming object detection. Since it does not deal with one data sample at a time, it is robust against changing training data with noisy information or data. A batch of samples is selected relying on the AL exploration philosophy based on uncertainty and diversity sampling, and the sampled streaming data are iteratively partitioned into bins for the incremental SSL method. In Fig. 1, the confident dataset consists of the samples labeled with correct object locations and object classes. The streaming images are unlabeled samples.

A tentative sample set includes not only incorrectly labeled samples but also biased labels under imbalanced underlying distribution. The noisy samples should be handled differently from the confident samples since such samples do harmful effect and never create any contribution to build a high-performing object detector.

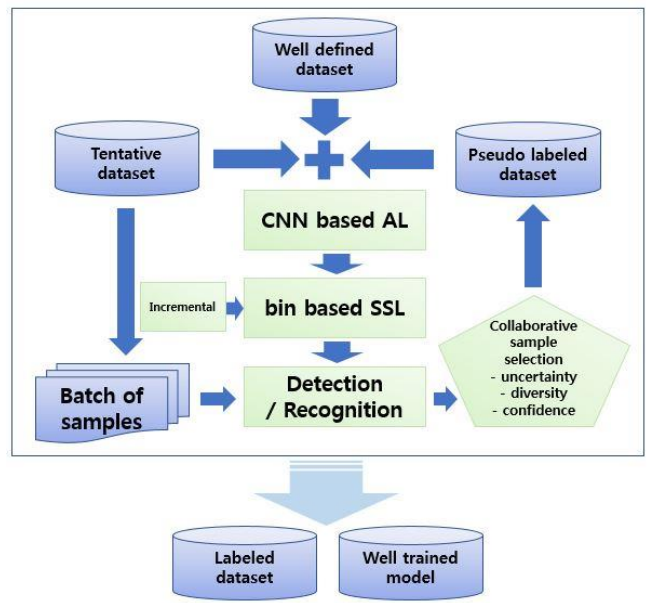

**FIGURE 1. The proposed IASSL framework, which consists of the nested learning cycle: that have the AL cycle for adaptive deep feature learning, and the incremental SSL cycle for bin-based learning.**

IASSL is initialized with a pre-trained CNN network and a confident, labeled dataset. The streaming image samples are filtered by collaborative sampling selection, which consists of the uncertainty and diversity criteria for AL and the confidence criterion for SSL. The collaboration between SSL and AL allows obtaining more confident and informative labeled training samples even from the noisy and unlabeled streaming samples. The selected samples show higher uncertainty in the true classes, and have lower confidence than the remaining samples. Diversity criterion is applied to samples to construct efficient batches of samples to incrementally improve the detection accuracy in each iteration. Learning with IASSL is divided into AL, using the confident dataset, and incremental SSL using streaming samples divided into bin sequences. In the AL cycle, a batch of object samples is divided into several bins, and the bin cycles are conducted for bin-based incremental SSL using the current confident dataset. IASSL initially trains the CNN using initial, confidently labeled samples, and it repeatedly retrains the next deep model for the CNN by adding the batch of samples selected using the current object detector until a convergence criterion is reached. In the SSL cycle, local streaming image samples are clustered, filtered by the collaborative sample selection, and the bin-based incremental learning is applied.

## IV. HIERARCHICAL OBJECT DETECTION TREE

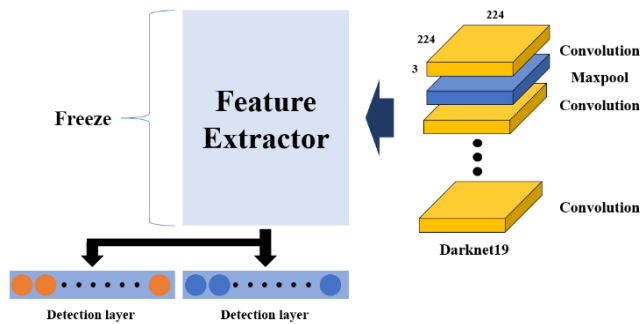

**FIGURE 2.** The flowchart for feature extraction from deep CNN architecture Darknet-19.

Incremental deep learning is based on the deep learning architecture introduced by Darknet-19 [11]. However, we introduce a hierarchical structure in addition to Darknet-19. We use mostly $3 \times 3$ filters, and every pooling step, doubled the number of channels, as shown in Fig. 2. We use $1 \times 1$ filters to compress feature representation and batch normalization to train the model. It has five max pooling layers and 19 convolutional layers.

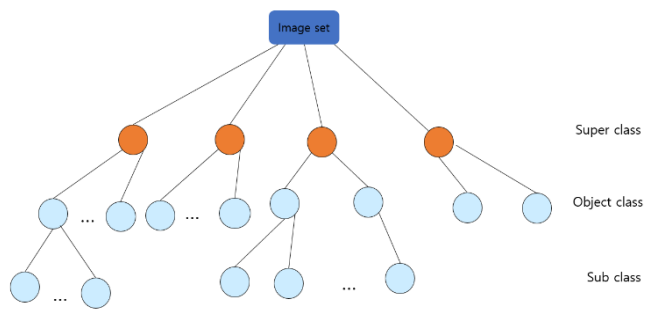

**FIGURE 3.** Hierarchical object detection tree using IASSL.

Hierarchical object detection based on IASSL is divided into three levels, as shown in Fig. 3. The three levels are the super class, object class, and sub class.

In Fig. 3, the top of the hierarchical tree represents the root, which has commonly shared convolutional layers to learn the common deep representations for all the object classes. In the root, the super class detector (i.e., indoor objects, animals, persons, and vehicles) is built using the convolutional layers and max pooling layers to learn the super classes. The first-level nodes of the hierarchical tree are associated with the object class detectors. The second-level nodes of the hierarchical tree are associated with the sub class detectors. This level can be expanded to represent unseen objects and new classes of objects. The third level of the hierarchical tree is replaced with the softmax layer in order to train deep CNN features jointly, since each group of the super class may contain a small number of existing object classes.

## V. INCREMENTAL DEEP LEARNING ALGORITHM

An object detection system predicts object class and location in terms of the conditional label and location distribution, $P(y, b|x)$, where $x$, $y$, and $b$ are feature vector, object label, and location, respectively. Even though the training sample distribution and the test sample distribution share the same conditional label probability distribution, prediction $P(y, b|x)$ is vulnerable to sample selection bias. In a streaming object detection system, a sampling method includes some sophisticated mechanisms to minimize the effects of sampling bias and class imbalance for acceptable detection accuracy.

### A. PROBLEM FORMALIZATION

In an incremental SSL step, the super class is decided by CNN detector $(C_1, C_2, ..., C_{sup}, D_{sup})$, which is constrained by the super class prototype model. Here $C_1$, $C_2$, and $C_{sup}$, represent first, second, and final super class convolution layer and $D_{sup}$ is super class detector layer. The object class detector is associated with a super class node in the first level of the hierarchical tree. Each super class node builds CNN detector $(C_1, C_2, ..., C_{obj}, D_{obj})$ to decide the object node by predicting the object class and bounding box. Here $C_1$, $C_2$, and $C_{obj}$, represent first, second, and final object class convolution layer and $D_{obj}$ is object class detector layer. The bounding box of feature vector $FV$, denoted by $BB(FV)$, indicates the bounding box region within which the best position of deep feature descriptor $FV$ can be found with high probability. The hierarchical tree has three types of object node based on confusion table analysis.

We analyzed each object class performance and categorized them based on the following three criteria.

i) Case 1: Existing object class, e.g., Inha University, Hi-tech Building table. If the class already exists in a benchmark dataset (e.g., PASCAL VOC) we consider that class an existing object class.

ii) Case 2: Combined object class with one or more existing object classes, for example, Inha hi-tech lobby, sofa of Inha University with PASCAL VOC sofa dataset. Because the class has a strong likelihood, and data size is not sufficient for a IASSL experiment, we combined the new object class data with the existing object class.

iii) Case 3: Local new object class as a new class, e.g., arena chair. If the new class does not have a single-likelihood value greater than the threshold, we consider that object class a new class.

Given a test scene image, $I$, and a training dataset, $\mathcal{D}$, assume that a prior distribution over $FV$ exists. Then, $FV$ can be treated as a random variable in Bayesian statistics. The posterior distribution of $FV$ is represented by

$$p(FV|I,\mathcal{D}) = \frac{p(I|FV,\mathcal{D})p(FV|\mathcal{D})}{\int} \qquad (1)$$

Since the hierarchical tree detector consists of super-class, object-class, and sub-class detectors, equation (1) can be rewritten as

$$p(FV_{sup}, FV_{obj}, FV_{sub}|I,\mathcal{D})$$
$$= p(FV_{sup}|I,\mathcal{D})p(FV_{obj}|FV_{sup},I,\mathcal{D})p(FV_{sub}|FV_{obj},FV_{sup},I,\mathcal{D})$$
$$= \frac{p(I|FV_{sup},\mathcal{D})p(FV_{sup},\mathcal{D})}{\int}$$
$$\frac{p(I|FV_{obj},FV_{sup},\mathcal{D})p(FV_{obj}|FV_{sup},\mathcal{D})}{\int}$$
$$\frac{p(I|FV_{sub},FV_{obj},FV_{sup},\mathcal{D})p(FV_{sub}|FV_{obj},FV_{sup},\mathcal{D})}{\int} \qquad (2)$$

where $FV_{sup}, FV_{obj}$, and $FV_{sub}$ are localized feature descriptions of super-class, object-class, and sub-class nodes, respectively. The maximum posterior of $(FV_{sup}, FV_{obj}, FV_{sub})$ is calculated as follows:

$$(\widetilde{FV}_{sup}, \widetilde{FV}_{obj}, \widetilde{FV}_{sub})_{MAP}(I,\mathcal{D})$$
$$= \underset{FV_{sup}, FV_{obj}, FV_{sub}}{\mathrm{argmax}} \begin{array}{l} [p(I|FV_{sup},\mathcal{D})p(FV_{sup},\mathcal{D}) \\ p(I|FV_{obj},FV_{sup},\mathcal{D})p(FV_{obj}|FV_{sup},\mathcal{D}) \\ p(I|FV_{sub},FV_{obj},FV_{sup},\mathcal{D})p(FV_{sub}|FV_{obj},FV_{sup},\mathcal{D})] \end{array} \qquad (3)$$

Since the object class is constrained by part-location search areas, $part_r$ is assumed to be treated as conditionally independent. Given priors, the hierarchical tree detector ensemble is looking for optimal feature vectors $FV_{sup}, FV_{obj}$, and $FV_{sub}$ satisfying the following:

$$(\widetilde{FV}_{sup}, \widetilde{FV}_{obj}, \widetilde{part}_1,\ldots,\widetilde{part}_R)_{MAP}(I,\mathcal{D})$$
$$= \underset{FV_{sup}, FV_{obj}, FV_{sub}}{\mathrm{argmax}} \begin{array}{l} p(I|FV_{sup},\mathcal{D})p(FV_{sup},\mathcal{D}) \\ p(I|FV_{obj},FV_{sup},\mathcal{D})p(FV_{obj}|FV_{sup},\mathcal{D}) \end{array} \prod_{r=1}^{R} p(I|part_r,FV_{obj},FV_{sup},\mathcal{D})p(part_r|FV_{obj},FV_{sup},\mathcal{D}) \qquad (4)$$

Note that $p(I|FV_{sup},\mathcal{D})$ and $p(I|FV_{obj},FV_{sup},\mathcal{D})$ are the likelihood functions of $FV_{sup}$ and are estimated by the super-class detector and object-class detector discussed in the following subsections. We minimize the negative of the logarithm of the posterior, rather than maximizing equation (5), as follows:

$$(\widetilde{FV}_{sup}, \widetilde{FV}_{obj}, \widetilde{part}_1,\ldots,\widetilde{part}_R)_{MAP}(I,\mathcal{D})$$

$$= \underset{FV_{sup}, FV_{obj}, FV_{sub}}{\mathrm{argmin}} \begin{array}{l} \log p(I|FV_{sup},\mathcal{D}) + \log p(FV_{sup},\mathcal{D}) \\ + \log p(I|FV_{obj},FV_{sup},\mathcal{D}) + \log p(FV_{obj}|FV_{sup},\mathcal{D}) \\ + \sum_{r=1}^{R} [\log p(I|part_r,FV_{obj},FV_{sup},\mathcal{D}) \\ + \log p(part_r|FV_{obj},FV_{sup},\mathcal{D})] \end{array} \qquad (5)$$

### B. FEATURE VECTOR OPTIMIZATION

We rewrite equation (5) by the block coordinate descent-based optimization formulation [40] as follows:

$$(\widetilde{FV}_{sup}, \widetilde{FV}_{obj}|I,\mathcal{D}) = \underset{FV_{sup}}{\mathrm{argmin}} f(FV_{sup}; FV_{obj}; FV_{sub}, I,\mathcal{D})$$
$$+ \underset{FV_{obj}}{\mathrm{argmin}} f(FV_{obj}; FV_{sup}; FV_{sub}, I,\mathcal{D}) \qquad (6)$$

The solution to equation (6) is a nonconvex, but it is a convex w.r.t. each of the optimization variables. We adopted an optimal algorithm based on the block coordinate descent method [40], which minimizes equation (6) iteratively w.r.t. each variable, while the remaining variables are fixed. The entire process is summarized in Algorithm 1.

| **Algorithm 1 Block Coordinate Descent-based Optimization** |
|---|
| **Input**: Image $I$, $\mathcal{D}$ with the hierarchical tree, regularization thresholds $(\lambda_1, \lambda_2)$ |
| **Output**: $(\widetilde{FV}_{sup}, \widetilde{FV}_{obj}, \widetilde{FV}_{sub})$ |
|   1. **Method:** |
|   2. Initialize $\widetilde{FV}_{sup}, \widetilde{FV}_{obj}$, and $\widetilde{FV}_{sub}$ |
|   3. $k \leftarrow 0$ |
|   4. **repeat** |
|   5. $\quad FV_{sup}^{k+1} \Leftarrow \underset{B}{\mathrm{argmin}} J_{sup}(FV_{sup}^k; FV_{obj}^k, FV_{sub}^k)$ |
|   6. $\quad FV_{obj}^{k+1} \Leftarrow \underset{C}{\mathrm{argmin}} J_{obj}(FV_{obj}^k; FV_{sup}^k, FV_{sub}^k)$ |
|   7. $\quad FV_{sub}^{k+1} \Leftarrow \underset{D}{\mathrm{argmin}} J_{sub}(FV_{sub}^k; FV_{sup}^k, FV_{obj}^k)$ |
|   8. **until** convergence. |

where $J_{sup}$, $J_{obj}$, and $J_{sub}$ are objective functions defined from equation (6), respectively. $\boldsymbol{B}$, $\boldsymbol{C}$, and $\boldsymbol{D}$ are optimization parameters for feature vectors $FV_{sup}$, $FV_{obj}$, and $FV_{sub}$.

Nevertheless, the detection result with the maximum likelihood may not be correct object detection, and it may not be consistent with other feature points. These kinds of errors always occur in most object detectors relying on a limited labeled training dataset. In many cases, detectors are unstable due to noise, pose variances, cluttered background, and illumination changes. The hierarchical object detection tree often under-fits due to the shortage of initial labeled training data; the class models are biased, and the classification boundaries determined by the hierarchical tree are often far from being the best choice. In this context, we introduce a data-driven semi-supervised framework that can learn incrementally using both labeled and unlabeled datasets to minimize the effects of troublesome patterns by preventing outliers.

## C. INCREMENTAL ACTIVE SEMI-SUPERVISED LEARNING

The proposed IASSL combines the uncertainty and diversity properties from the AL paradigm and the confidence property from the incremental SSL paradigm. It minimizes not only the training time but also costly human intervention, and at the same time, keeps a high-quality training dataset similar to [27]. Considering the uncertainty criterion of AL, the most uncertain samples are selected as the most useful training samples to be added, since those are expected to be incorrectly classified by the current classification model with high probability. However, the uncertainty criterion may cause the selection of noisy or redundant samples. We adapted a pool-based (batch or bin) AL learning framework combined with an incremental SSL philosophy, based on the collaborative sampling method of AL and SSL in terms of uncertainty, confidence, and diversity criteria, which are expected to select more informative and training samples with low redundancy.

We use an AL batch cycle similar to our previous work [36] and added a bin cycle for incremental SSL. In the AL batch cycle, a training dataset is divided into well-defined labeled training samples, $D_{well}$, and weakly or unlabeled training samples, $D_{tentative}$. IASSL processes them to increase the volume of $D_{well}$ above that of $D_{tentative}$. Initially, the IASSL-based image processing system learns using the well-defined dataset, which is used to construct the pre-trained CNN. This dataset is assumed to be correctly pre-labeled. The initial well-defined dataset is sampled from the data that are used to construct the CNN's pre-trained model. A batch of samples is selected considering the distribution of the prototype models and class balancing. The confidence scores are (re)assigned to the weak samples by the current (CNN) detector. Confident and well-defined samples are selected from the weak samples according to the confidence score measured and ranked by the current (CNN) detector. A subset of weak samples is selected using the collaborative sampling strategy, whereby the current detector reassigns new labels or assign labels with high scores; some ambiguous samples will be removed or relabeled by oracle after being filtered by uncertainty and diversity criteria. Note that we can minimize human effort by exploring only a small portion of the weakly labeled samples, while classification accuracy can be improved. More informative samples reflecting the diversity criterion of the active learning paradigm are mined for a current batch of samples. The selected confident samples are added to the current training dataset and generate a new well-defined training dataset to retrain the CNN detector.

Instead of only the selection of one sample in an iteration, IASSL minimizes the learning time by building a pool of $D_\Delta$ samples based on the uncertainty, diversity and confidence criterion, where $D_\Delta$ is a final sample set determined at operation time, and a sample pool can be batch or bin (see Fig. 1). For each class, a set of samples is selected and scored using the current detector and added to the training set. We select $\eta$ samples with closer object class scores, $f(x)$, in each half of the margin according to the uncertainty principle. We have a total of $\eta$ samples for what we call candidate samples. From the total we select $\eta$ samples with a $f(x)$ score that is between 0 and 1. When uncertainty parameter $\eta$ decreases then $f(x)$ distance increase.

However, the uncertainty criterion cannot avoid the selection of similar samples. IASSL also provides the advantage of being incorporated with a diversity measure [33]. The candidate pool of samples with a diversity criterion is determined by selecting $\vartheta$ samples from the $\eta$ candidates with a more diverse property, similar to [27]. The distribution of the remaining samples is analyzed by K-means clustering algorithm to determine the uncertainty criterion. IASSL evaluates the distribution of the selected $\eta$ samples based on standard k-means clustering and removes outliers and similar samples. We define candidate sample set $D_{diversity}$, which contains more informative samples within the rank $\vartheta$ measured by the deep confident scores, i.e., $f(x)$:

$$D_{diversity} = \{X | X \in D_{tentative}, 0 \le f(X) \le 1\}$$
$$\text{s.t. } Rank(x) < \eta \qquad (7)$$

where $Rank(x)$ denotes the decreasing order of $f(x)$ measured by the class score of the current object detector. We have constructed the batches of samples by incorporating a diversity measure.

Next, we applied the incremental SSL philosophy by initializing $D_\Delta$ with the sample $X_{top} = argmax_{x \in D_{diversity}} f(x)$, $X_{top} \in D_{diversity}$ using confidence criterion parameter γ. At each step, our sampling strategy chooses a sample from $D_{diversity}$ and adds to $D_\Delta$. $D_{diversity}$ becomes the most similar sample in $D_\Delta$ in terms of confidence score, i.e.,

$$X_{top} = argmax_{x \in D_{diversity}}\{max_{x_i, x_j \in D_\Delta} d(x_i, x_j)\} \qquad (8)$$

In (8), we use Euclidian distance between two features to calculate $d(x_i, x_j)$.

When the cardinality of $D_\Delta$ becomes γ, the sample selection process is stopped, and the final sample set is $D_\Delta$. We retrain the CNN using the pool of samples, and the process is repeated until a convergence criterion is satisfied. The entire process is summarized in Algorithm 2.

---

**Algorithm 2 Batch-Bin-cycle Active Semi-Supervised Learning Using Collaborative Sampling Selection**

**Input:** Well labeled dataset $D_{well}$ and tentatively labeled dataset $D_{tentative}$ with $D_{well} \ll D_{tentative}$.
Notations:
$D_\Delta$, batch dataset
$D_{tentative}$, batch tentative dataset with $D_\Delta \ll D_{tentative}$.

$\eta$, uncertainty parameter
$\vartheta$, diversity parameter
$\gamma$, confidence parameter
$f$, detector
$f(x)$, class score by detector
$Rank(x)$, decreasing order by $f(x)$
$Acc$, accuracy of bin
**Output:** Expanded confident labeled dataset $D_{well}$, an optimal object detector $f$

1. **Step 1**: Train the initial CNN detector $f$ using $D_{well}$.
2. **Repeat Step 2 to Step 6** until convergence
3. **Step 2:** Select a batch pool of candidate samples $D_\Delta$ from $D_{tentative}$. Compute $f(x)$, $x \in D_\Delta$ using $f$.
4. **Step 3:** Create a batch dataset as follows.
5. 3-1. Select $\eta$, $\vartheta$ tentatively labeled samples filtered by uncertainty and diversity criteria. $D_{diversity} = \{X | X \in D_{tentativ}\}$ s.t. $Rank(x = \vartheta < \eta)$
6. 3-2. Confidence sampling for incremental SSL is initialized by the sample $X_{top} = argmax_{x \in D_{diversity}} \{max_{x_i, x_j \in D_\Delta} d(x_i, x_j)\}$
7. 3-3. Repeat until $|D_\Delta| = \gamma$; $D_\Delta = D_\Delta \cup \{X_{top}\}$
8. **Step 4:** Assign pseudo-label and score to each unlabeled $D_\Delta$
9. **Step 5:** For each subspace
10. 5-1. Sort the pseudo-labeled tentative samples $D_\Delta$ in decreasing order.
11. 5-2. Split j bins sorted tentative samples in decreasing order such that $i^{th}$ bin has samples in the range $(i-1)/|D_\Delta|$ to $i/|D_\Delta|$. Generate bin sequence $BSeq = [bin_i]_{i=0}^j$ by partitioning $D_\Delta$.
12. **Step 6:** Repeat until $BSeq(i) \neq \phi$
13. 6-1. For each bin, $bin_i = BSeq(i)$, train $f^{(i+1)}$ using $D_{train}^{(i)} \cup bin_i$ and calculate $Acc_{bin_i}^{(i)}$.
14. 6-2. $Acc^{(i+1)} = \max_{bin_i}\{Acc_{bin_i}^{(i+1)}\}$
15. 6-3.
    If $Acc^{(i+1)} \geq Acc^{(i)}$, $bin^* = \underset{bin_i}{argmax}\{Acc^{(i+1)}\}$;
    $D_{train}^{(i+1)} = D_{train}^{(i)} \cup bin^*$; $f^{(i+1)} = f_{bin^*}^{(i+1)}$
    Else if $Acc^{(i+1)} < Acc^{(i)}$, oracle labels incorrectly labeled data in $bin^*$ and return $f^{(i+1)} = f^{(i)}$, $i++$
16. **Step 7:** Retrain $f$ using $D_{well} = D_{well} \cup D_\Delta$. $D_{tentative} = D_{tentative} - D_\Delta$;

## VI. EXPERIMENTS

The main goal of experiment is to confirm the efficiency of IASSL framework. To achieve this goal, we conducted a number of experiments on benchmark datasets, such as PASCAL VOC as well as a local dataset, and compared the results with state-of-the-art detectors, such as YOLOv2. All implementations are on a single server with cuDNN [41], a single NVIDIA TITAN X and Tensorflow [42].

### A. DATASET OVERVIEW

#### PASCAL VOC DATASET

The PASCAL VOC 2007 dataset [16] contains 20 classes with four categories: person, animal, vehicle, indoor. Among them, our experimental subject is the indoor data category, which includes bottle, chair, dining table, potted plant, sofa, and tv/monitor. It has a total of 9963 images (train/validation/test) with 24,640 annotated objects. On the other hand, PASCAL VOC 2012 also has 20 classes with 11,530 images (train/validation/test) containing 27,450 annotated objects. The YOLOv2 VOC model was trained with the PASCAL VOC 2007 trainval dataset and the PASCAL VOC 2012 trainval dataset.

#### LOCAL DATASET

In our experiment, we used a chair dataset of 400 chair images recorded in Gangneung Ice Arena, Korea. We also collected 400 sofa images and 400 table images from the Hi-tech building lobby of the Inha University campus.
From these 400 chair images, 10 chair images were considered for the initial dataset, 90 images for the test dataset, and the rest of the 300 images were considered for an unlabeled dataset. This dataset does not provide good detection results with YOLOv2, which is state-of-the-art object detection technology [10], trained with the existing PASCAL VOC dataset.

To train the dataset, our experiment settings are as follows: we used the Darknet-19 classification model [10] where the base detector is YOLOv2, and input image resolution is $416 \times 416$ pixels.

**IEEE** Access
Multidisciplinary | Rapid Review | Open Access Journal

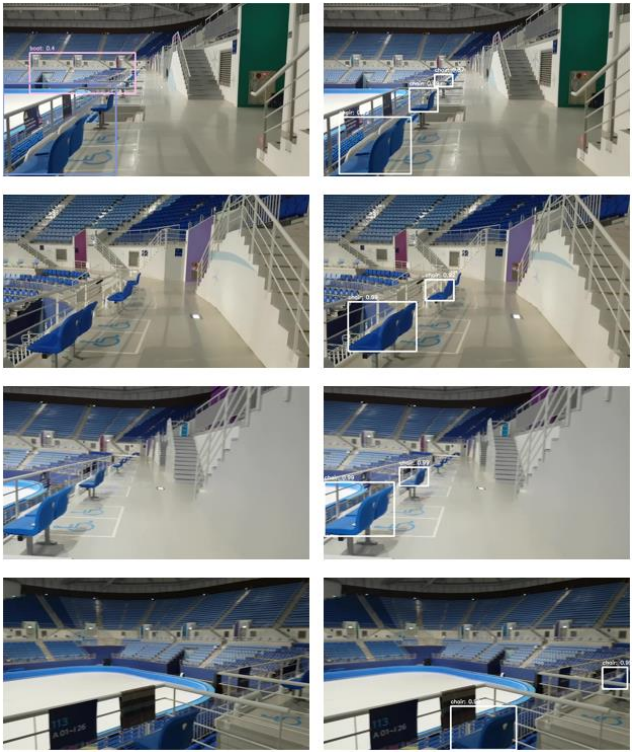

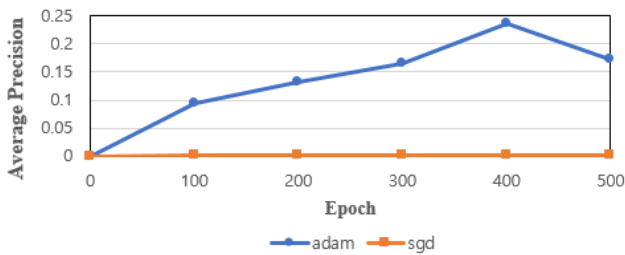

### B. EXPERIMENT PARAMETER SETTINGS

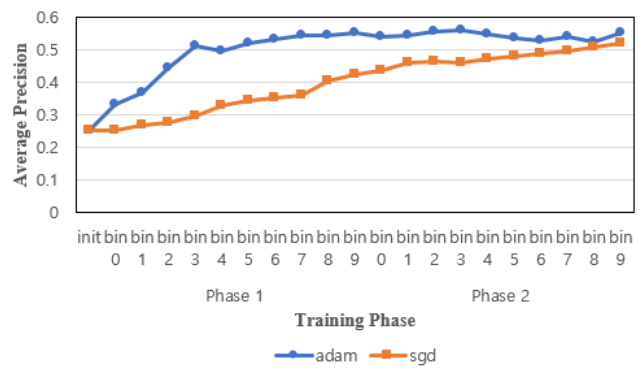

**FIGURE 5.** Adam and SGD optimizer performance on the chair dataset.

Average precision is used for accuracy in object detection. So, these figures show average precision for accuracy on the test data. We selected the gradient-based optimization method as Adam by the following reasoning at the beginning stage of our experiment. In Fig. 5, the y-axis shows the average precision (AP), and the x-axis shows the number of epochs. In this place, training results of the last layer use stochastic gradient descent (SGD) [43] and the Adam optimizer [44] to create the initial model. Here, SGD optimizer training did not work properly because it is very slow, whereas Adam optimizer training continues up to 500 epochs, and the learning rate is 0.001.

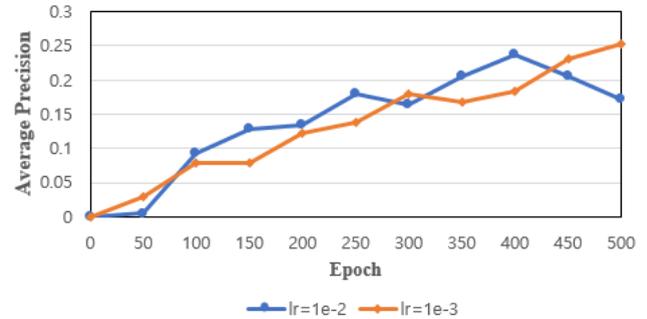

**FIGURE 6.** Adam optimizer training performance on the chair dataset. Here, lr means learning rates at 0.01 and 0.001 with 500 epochs.

Fig. 6 shows the different experiment results using different learning rates. According to the experiments, training results of the initial model were obtained by the last layer, where the comparison is completed using the Adam optimizer, and the learning rates are 0.01 and 0.001. We selected a learning rate of 0.001, which shows higher and stable performance.

**FIGURE 7.** IASSL bin training in phase 1 and phase 2.

In Fig. 7, we divided the training steps into phase 1 and phase 2. The total number of bins is 10 in each phase. Both SGD and Adam optimizer were tested for IASSL bin training in our experiment. We found that the Adam optimizer gets faster (time) convergence than SGD with a higher AP. The above experiments led us to use the Adam optimizer with a learning rate of 0.001 in the following experiments.

### C. EXAMPLE OF THE HIERARCHICAL OBJECT DETECTION TREE

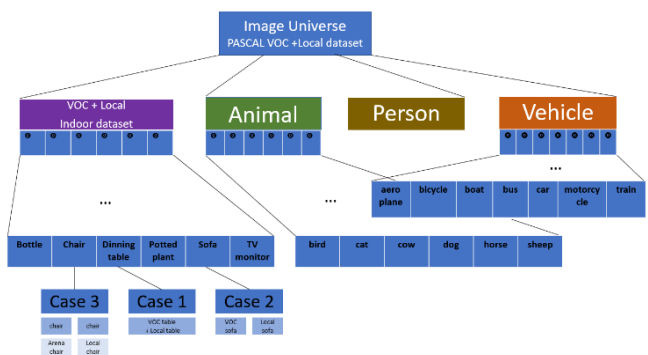

**FIGURE 8.** Data model for IASSL: the root node consists of PASCAL VOC 2007 and 2012 and the local dataset. The local datasets are captured.

Fig. 8 shows the hierarchical object detection tree, which consists of both PASCAL VOC data and local data. Although, the dataset contains a number of different data objects (animal, person, and vehicle), our work mainly focused on the indoor dataset, specifically, chair, sofa, and table images, which have three different cases (case-1, case-2, and case-3).

### D. EFFECT OF COLABORATIVE SAMPLING PARAMETERS

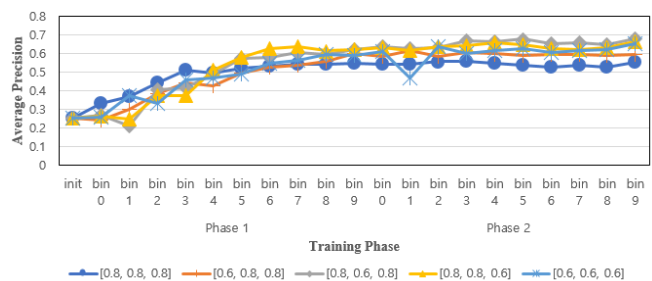

**FIGURE 9.** Results for AP in the test dataset by adjusting the IASSL collaborative sampling parameters. The parameter combination [a, b, c] indicates uncertainty, diversity, and confidence factors, respectively, in collaborative sampling.

In Fig. 9, we show the effect of different parameters for uncertainty, diversity, and confidence sampling methods. The best result is achieved using the parameter combination [0.8,0.6,0.8], i.e. a grey color. The second-best result is obtained using [0.8,0.8,0.6], i.e. a yellow color. Here, we use the Adam optimizer with a learning rate of 0.001.

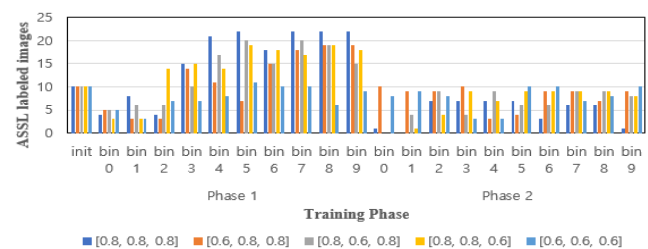

**FIGURE 10.** The number of images sampled and labeled by the IASSL method during each sampling phase.

Fig. 10 shows the whole labeled dataset after each phase. We can see that in phase 1 the parameter combination [0.8,0.8,0.8] in the dark blue color in Fig. 10, generates the highest number for the labeled dataset. However, considering the performance, we can select the parameter combination as either [0.8,0.6,0.8] or [0.8,0.8,0.6].

We investigated the effect from increasing the number of IASSL phases to three from two. The experiment results shown in Fig. 10 use a learning rate of 0.001, and the number of IASSL phases was increased to three from two through sampling parameters [0.8,0.6,0.8] and [0.8,0.8,0.6]. In this experiment we set the parameter with learning rate 0.001 and use the Adam optimizer.

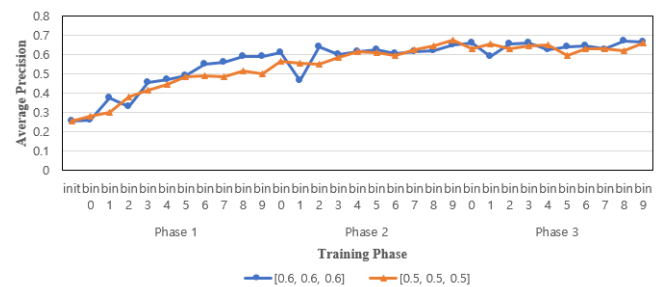

**FIGURE 11.** Training sample images up to phase 3.

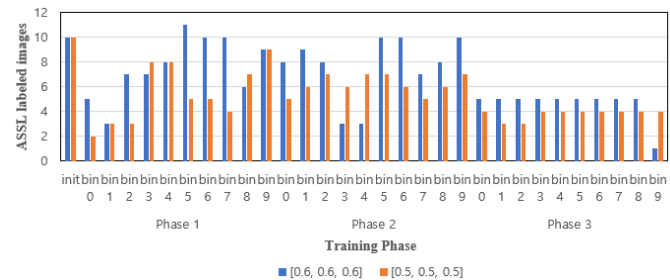

**FIGURE 12.** The number of images sampled and labeled during the experiment in Fig. 11.

In Fig. 11. and Fig. 12. show that phase 2 and phase 3 do not differ greatly in performance, even when [0.6, 0.6, 0.6] converges faster than [0.5, 0.5, 0.5] and progresses to phase 3.

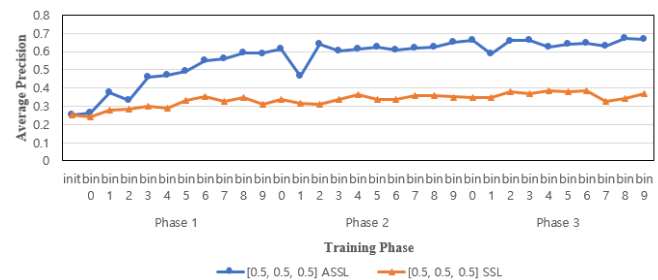

**FIGURE 13.** ASSL and SSL performance comparison.

The ASSL and SSL performance results are shown in Fig. 13. We considered a fixed number of sampling parameters, [0.5, 0.5, 0.5] for uncertainty, diversity, and confidence, for the ASSL and SSL experiment. While ASSL has greatly improved performance, SSL performance merely improved or showed no change at all. We use the Adam optimizer and set the learning rate of 0.001 in above experiment.

### E. Comparison with the collaborative sampling with other sampling method

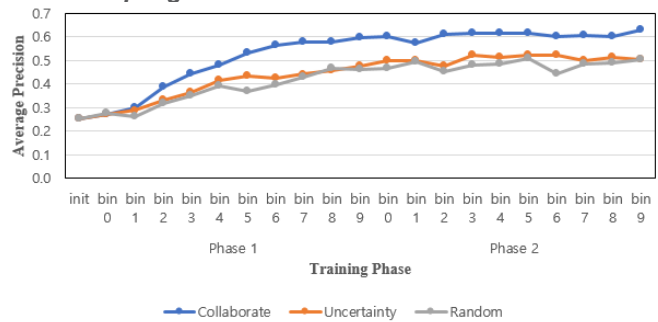

**FIGURE 14.** Result of collaborate sampling and other methods

Uncertainty sampling is one of the popular sampling method of active learning [30]. This sampling method used in many state-of-the-art techniques where active learning work as catalyst for object detection [45], [46]. In Fig. 14, we show the comparison result of our collaborative sampling, uncertainty sampling, and random sampling methods. In our experiment, collaborative sampling method show higher mAP than uncertainty, random sampling methods. We set the learning rate 0.001 and use the Adam optimizer in this experiment.

### F. COMPARISON WITH YOLOv2

In our experiment, a popular object detector is introduced and compared with IASSL. This object detector was trained with both a benchmark dataset and our local dataset for a fair evaluation.

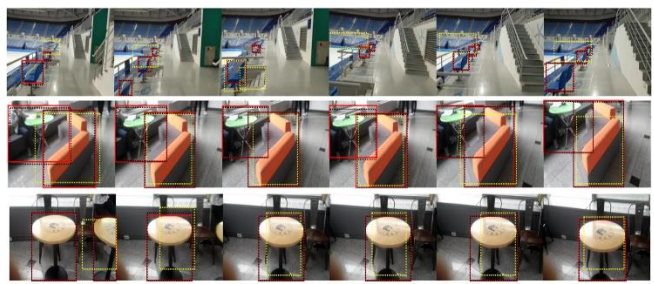

**FIGURE 15.** The detection result from IASSL on local datasets of chairs, sofas, and tables.

The detection results for chairs, sofas, and tables with the local dataset are shown in Fig. 15. In all cases, the red bounding box represents ground truth, the black bounding box indicates IASSL, and the yellow bounding box represents the YOLOv2 VOC model. Each of these objects has 100 labeled data items and 300 unlabeled data items. Thus, local chair, sofa, and table images were combined with the PASCAL VOC test dataset for a fair evaluation. Here, the black bounding box shows outstanding performance when detecting chairs, sofas, and tables. Besides, IASSL performs well under various illumination changes. Here the Adam optimizer used with a learning rate of 0.001 for the parameter selection.

**TABLE 1.** Comparison results of state-of-the-art object detection algorithms and our method

| Method | $(0,100)^*$ | (50,50) | (70,30) | (80,20) | (90,10) |
|---|---|---|---|---|---|
| Faster RCNN | 71.1 | 71.7 | 72.3 | 72.6 | 72.8 |
| SSD300 | 71.8 | 72.8 | 73.3 | 73.6 | 73.8 |
| YOLOv2 | 74.5 | 74.7 | 75.1 | 75.1 | 75.1 |
| IASSL** (0.8, 0.8, 0.8) | 77.2 | 76.3 | 76.3 | 76.0 | 75.9 |
| IASSL (0.6, 0.8, 0.8) | 77.5 | 75.8 | 75.9 | 75.7 | 75.6 |
| IASSL (0.8, 0.6, 0.8) | **77.8** | **76.6** | **76.6** | **76.4** | **76.2** |
| IASSL (0.8, 0.8, 0.6) | 77.7 | 76.0 | 76.0 | 75.9 | 75.9 |
| IASSL (0.6, 0.6, 0.6) | 77.7 | 75.9 | 76.0 | 75.8 | 75.7 |
| IASSL (0.5, 0.5, 0.5) | 77.8 | 76.0 | 76.1 | 75.9 | 75.8 |

*In the first row (a, b) means the test data are combined at a% from the VOC test data and b% from Inha local data.
**IASSL (u, d, c) indicates the collaboration parameters, i.e., uncertainty, diversity, and confidence.

Table 1 shows the comparison results of state-of-the-art object detection algorithms and our method. Here, we used the VOC 2007 test data and our local data; (a, b) is the composition ratio for all data, in which "a" represents VOC 2007 test data, and "b" represents the ratio of local data; mAP is compared for each test dataset. In the IASSL method (u, d, c) are uncertainty, diversity, and confidence, respectively. We can see that the proposed method outperforms the famous object detectors: faster RCNN, SSD300, and YOLOv2.

**TABLE 2.** Comparison results of state-of-the-art object detection algorithms and our method in COCO validation 2017 dataset

| Method | $(0,100)^*$ | (50,50) | (70,30) | (80,20) | (90,10) |
|---|---|---|---|---|---|
| YOLOv2 | 32.58 | 32.38 | 32.13 | 31.81 | 31.42 |
| IASSL** (0.8, 0.8, 0.8) | 32.94 | 32.74 | 32.44 | 32.14 | 31.74 |
| IASSL (0.6, 0.8, 0.8) | 32.97 | 32.77 | 32.47 | 32.17 | 31.77 |
| IASSL (0.8, 0.6, 0.8) | **33.02** | **32.82** | **32.52** | **32.22** | **31.82** |
| IASSL (0.8, 0.8, 0.6) | 33.02 | 32.81 | 32.51 | 32.21 | 31.81 |
| IASSL (0.6, 0.6, 0.6) | 33.01 | 32.81 | 32.51 | 32.21 | 31.81 |
| IASSL (0.5, 0.5, 0.5) | 33.01 | 32.81 | 32.51 | 32.21 | 31.81 |

*In the first row (a, b) means the test data are combined at a% from the COCO validation data and b% from Inha local data.
**IASSL (u, d, c) indicates the collaboration parameters, i.e., uncertainty, diversity, and confidence.

Table 2 shows the comparison results of YOLOv2 and our method. Here, we used the COCO 2017 validation data [17] and our local data; (a, b) is the composition ratio for all data, in which "a" represents COCO 2017 validation data, and "b" represents the ratio of local data; mAP is compared for each dataset. Our proposed method outperforms the YOLOv2.

Some of the previous work in literature considered incremental learning where they used object detector such as Fast RCNN and Faster RCNN [47], [48]. Similarly, our method included YOLOv2 which perform well on both local and benchmark dataset. We showed comparison results in Table 1 and Table 2.

### G. Experiment on noisy data

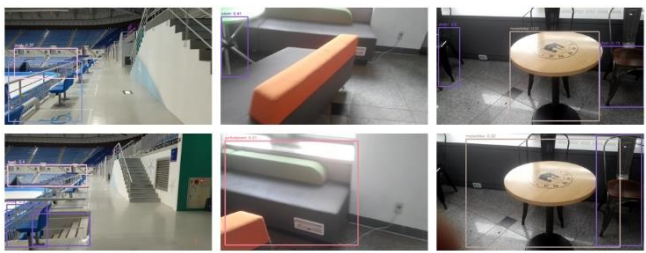

**FIGURE 16.** Examples of noisy labeled data

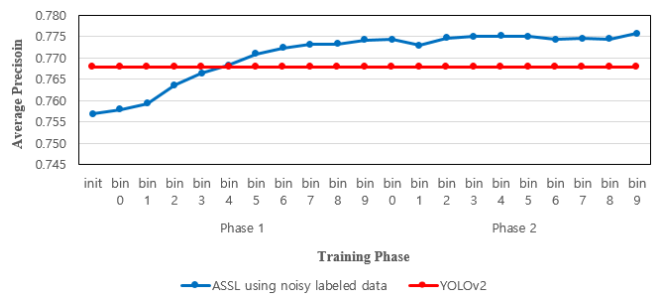

**FIGURE 17.** Result of ASSL using noisy labeled local data

Yuan *et al.* propose Iterative Cross learning (ICL) that significantly improve performance on noisy dataset, we applied ASSL technique on similar noisy labeled dataset in the experiment as shown in Fig. 16. and Fig 17 [49]. Here they applied ICL in image classification, on the other hand, we apply ASSL in object detection where the position of the bounding box and the class is labeled noisy too. Fig. 16. shows an example of noisy labeled data. We used the VOC 2007 test data and our local noisy labeled data. In Fig. 17, our ASSL method shows that at the beginning the mAP was low due to the noisy labeled data but after bin 4 in training phase 1 the mAP improved. We use the Adam optimizer with a learning rate of 0.001 in the above experiment.

## VII. CONCLUSION

In this paper, we proposed Incremental active semi-supervised learning combining batch-based active learning and bin-based semi-supervised learning using a collaborative sampling strategy to achieve high performance. This improves learning algorithms for a streaming object detector in the presence of changing and noisy environments. Active learning uses a collaborative sampling method for measurement of uncertainty and diversity, and it collaborates with semi-supervised learning based on a confidence criterion. Our proposed model produces higher performance with fewer errors, higher accuracy, and less human effort, in comparison with semi-supervised learning methods. These achievements encourage further improvement of the proposed method. Future research directions include finding ways to deal more flexibly with the learning sequence that reduces the number of erroneous labeled data and that of discarded good labeled data.

## ACKNOWLEDGMENT

This work was supported by Inha University research grant. The GPUs used in this research were generously donated by NVIDIA.

**IEEE** Access

Multidisciplinary : Rapid Review : Open Access Journal